\documentclass[conference]{IEEEtran}
\IEEEoverridecommandlockouts
\usepackage{cite}
\usepackage{amsmath,amssymb,amsfonts}
\usepackage{algorithmic}
\usepackage{graphicx}
\usepackage{textcomp}
\usepackage{xcolor}
\usepackage[caption=false]{subfig}
\usepackage{booktabs}
\usepackage{balance}

\def\BibTeX{{\rm B\kern-.05em{\sc i\kern-.025em b}\kern-.08em
    T\kern-.1667em\lower.7ex\hbox{E}\kern-.125emX}}

\usepackage{acronym}
\newacro{CAM}{Class Activation Mapping}
\newacro{CNN}{Convolutional Neural Network}
\newacro{AI}{Artificial Intelligence}
\newacro{CV}{Computer Vision}
\newacro{XAI}{Explainable Artificial Intelligence}
\newacro{DTD}{Deep Taylor Decomposition}
\newacro{LRP}{Layer-wise Relevance Propagation}
\newacro{ABN}{Attention Branch Network}
\newacro{SGD}{Stochastic Gradient Descent}
\newacro{IoU}{Intersection over Union}
\newacro{GAP}{Global Average Pooling}
\newacro{PCA}{Principal Component Analysis}
\newacro{DNN}{Deep Neural Network}
\newacro{DL}{Deep Learning}
\newacro{ML}{Machine Learning}
\newacro{NN}{Neural Network}
\newacro{VQI}{Visual Quality Inspection}
\newacro{MLOps}{Machine Learning Model Operationalization Management}
\newacro{COCO}{Common Objects in Context}
\newacro{RISE}{Randomized input sampling for explanation}
\newacro{ASPP}{Atrous Spatial Pyramid Pooling}
\newacro{SHAP}{SHapley Additive exPlanations}
\newacro{XIL}{eXplanatory Interactive Learning}
\newacro{RRR}{Right for the Right Reasons}
\newacro{TPU}{Tensor Processing Unit}
\newacro{FPGA}{Field Programmable Gate Array}
\newacro{FCN}{Fully Convolutional Network}
\newacro{HLS}{High-Level Synthesis}
\newacro{RNN}{Recurrent Neural Networks}
\begin{document}

\title{XAI-Enhanced Semantic Segmentation Models for Visual Quality Inspection}


\author{\IEEEauthorblockN{
Tobias Clement\IEEEauthorrefmark{1}\IEEEauthorrefmark{2},
Truong Thanh Hung Nguyen\IEEEauthorrefmark{1}\IEEEauthorrefmark{2}\IEEEauthorrefmark{4}\thanks{\IEEEauthorrefmark{1}Equal Contribution},
Mohamed Abdelaal\IEEEauthorrefmark{3},
Hung Cao\IEEEauthorrefmark{4}},
\IEEEauthorblockA{
\IEEEauthorrefmark{2}Friedrich-Alexander-University Erlangen-Nürnberg, Germany\\
\IEEEauthorrefmark{3}Software AG, Germany
\IEEEauthorrefmark{4}Analytics Everywhere Lab, University of New Brunswick, Canada\\
Email: \{tobias.clement, hung.tt.nguyen\}@fau.de, mohamed.abdelaal@softwareag.com, hcao3@unb.ca}
}

\maketitle

\begin{abstract}Visual quality inspection systems, crucial in sectors like manufacturing and logistics, employ computer vision and machine learning for precise, rapid defect detection. However, their unexplained nature can hinder trust, error identification, and system improvement. This paper presents a framework to bolster visual quality inspection by using CAM-based explanations to refine semantic segmentation models. Our approach consists of 1) Model Training, 2) XAI-based Model Explanation, 3) XAI Evaluation, and 4) Annotation Augmentation for Model Enhancement, informed by explanations and expert insights. Evaluations show XAI-enhanced models surpass original DeepLabv3-ResNet101 models, especially in intricate object segmentation. 

\end{abstract}


\section{Introduction}
\acf{VQI} systems use \ac{AI} for automated quality inspections, reducing human errors and enhancing efficiency. While \acp{DNN} have improved VQI accuracy, they often compromise interpretability~\cite{baryannis2019predicting}, creating challenges due to their ``black box" nature, especially in critical domains~\cite{nguyen2023towards}.

\acf{XAI} seeks to make AI decisions understandable to humans~\cite{clement2023xair}. Beyond enhancing trust, it aids in model debugging and ensures fairness and compliance~\cite{molnar2019}. However, a framework combining transparency, reliability, and fairness for VQI systems, particularly with semantic segmentation models, is lacking. To fill this void, we introduce an XAI-augmented VQI framework using CAM-based explanations to refine models like DeepLabv3-ResNet101. Our goal is to balance model accuracy with interpretability.

Our main contributions include:
\begin{enumerate}
    \item VQI Framework Enhancement (Section~\ref{sec1}): We present a framework merging XAI with VQI systems, encompassing model training, explanation, XAI assessment, and enhancement.
    \item CAM Explanation Assessment (Section~\ref{sec2}): We evaluate the reliability and credibility of CAM explanations, guiding the choice of XAI methods.
    \item XAI-driven Model Optimization (Section~\ref{sec3}): We refine the DeepLabv3-ResNet101 model using annotations informed by CAM explanations and expert insights.
\end{enumerate}

The paper's structure is: Section~\ref{sec:related_work} reviews related work on visual quality inspection, segmentation, and XAI. Section~\ref{sec1} details our VQI use case and the XAI-integrated framework. Section~\ref{sec:evaluation} discusses our findings, leading to conclusions in Section~\ref{sec:conclusion}.

\section{Background \& Prior Research}\label{sec:related_work}
%
This section delves into four pivotal domains relevant to our research: visual quality inspection, semantic segmentation, XAI, and XAI-driven model enhancement.

\textbf{Visual Quality Inspection:} Quality control, integral to manufacturing, can be expensive and lengthy~\cite{tang2017manufacturing}. \ac{VQI}, an AI innovation, offers a reliable and consistent alternative~\cite{sun2018research}, benefiting industries like automotive and electronics~\cite{sun2018research, md2022review, yasuda2022aircraft}. Advanced DL models, such as YOLO~\cite{redmon2016you} and ResNet~\cite{he2016deep}, have greatly improved VQI efficiency~\cite{sundaram2023artificial}.

\textbf{Semantic Segmentation:} Essential for VQI, semantic segmentation labels image pixels, allowing VQI systems to focus selectively~\cite{guo2018review}. Notable models include FCN~\cite{long2015fully}, DeepLabv3~\cite{deeplabv32018}. We employ DeepLabv3, optimized with ResNet101, known for its mobile-friendly performance and effective multi-scale contextual capture~\cite{he2016deep}.

\textbf{Explainable AI:} XAI tools in CV reveal the workings of deep CNN models. Classifications include Backpropagation-based, CAM-based, and Perturbation-based methods~\cite{rebuffi2020there}. However, the plethora of XAI techniques can overwhelm users~\cite{clement2023xair}. Evaluations, thus, are essential. Metrics to evaluate XAI include plausibility and faithfulness, which align explanations with human intuition and the model's logic, respectively~\cite{hedstrom2022quantus, nguyen2023towards}.

\textbf{Model Enhancement with Explainable AI:} XAI can bolster model robustness, efficiency, reasoning, and fairness~\cite{weber2022beyond}. Enhancement strategies using XAI encompass:
\begin{itemize}
\item \textit{Data augmentation}: Techniques, like Guided Zoom~\cite{bargal2018guided}, and synthetic samples, can refine predictions and enhance performance.
\item \textit{Feature augmentation}: Approaches such as relevance-based feature masking~\cite{schiller2019relevance} and feature transformations target essential features and bias removal.
\item \textit{Loss augmentation}: Techniques, like \ac{ABN}~\cite{fukui2019attention}, modify the loss function with insights from XAI for better performance and reasoning.
\item \textit{Gradient augmentation}: Methods like \ac{LRP}~\cite{ha2019improvement} enhance model performance by optimizing gradients.
\item \textit{Model augmentation}: Strategies such as pruning and knowledge transfer can streamline models or recreate them with improved attributes.
\end{itemize}

\section{Methodology}\label{sec1}
%

This section unfolds our strategy to craft an advanced VQI system, leveraging XAI for optimal performance and transparency.

\textbf{Use Case – Visual Quality Inspection:} Focusing on a cloud-based AI solution, we aid field engineers in photographing assets through mobile devices. The cloud AI system discerns the asset type and health. The results subsequently update an asset management system, assisting in planning maintenance and providing on-field insights. To address challenges like calibration and unexpected data variations~\cite{rovzanec2022active}, we propose integrating XAI for clear and interpretable AI decisions.

\textbf{Dataset:} We employ the TTPLA dataset, crucial for identifying power-grid assets~\cite{abdelfattah2020ttpla}. Comprising various image scenarios, it is ideal for detection and segmentation.

%
\begin{figure}[htp]
    \centering
    \subfloat[][]{\includegraphics[width=.24\linewidth]{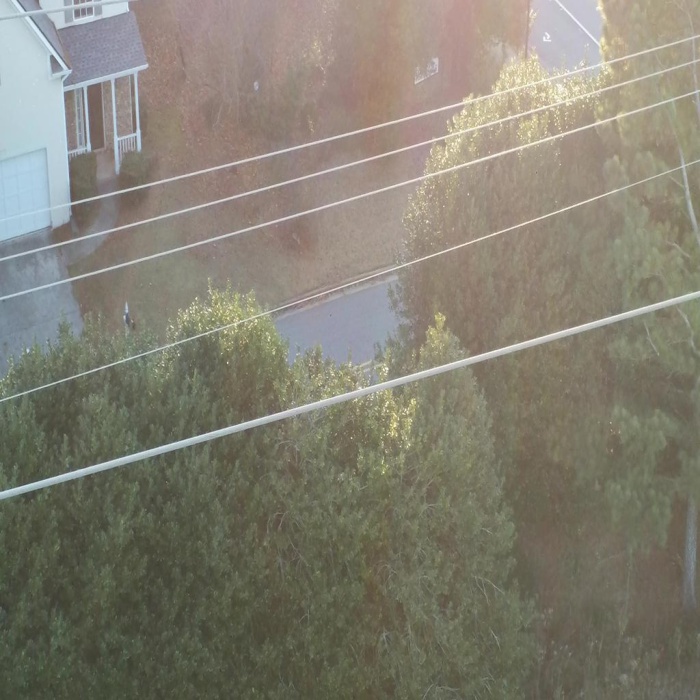}}\hfill
    \subfloat[][]{\includegraphics[width=.24\linewidth]{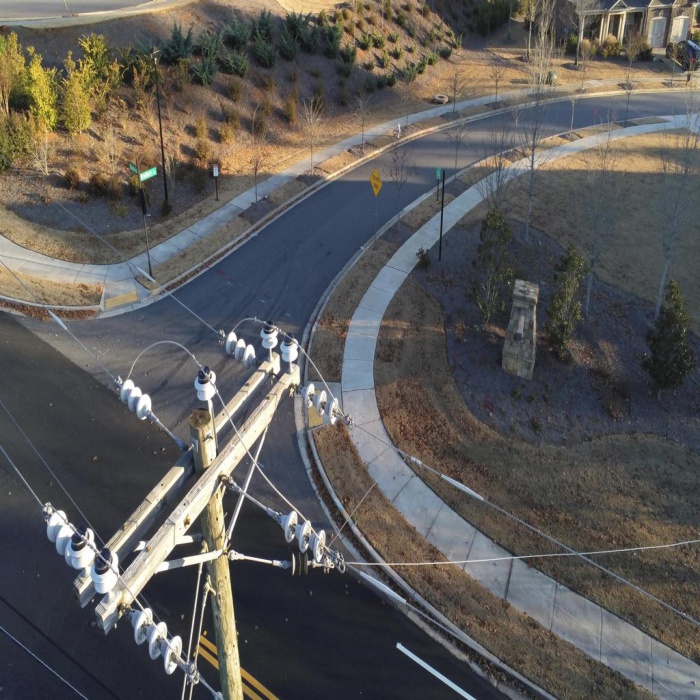}}\hfill
    \subfloat[][]{\includegraphics[width=.24\linewidth]{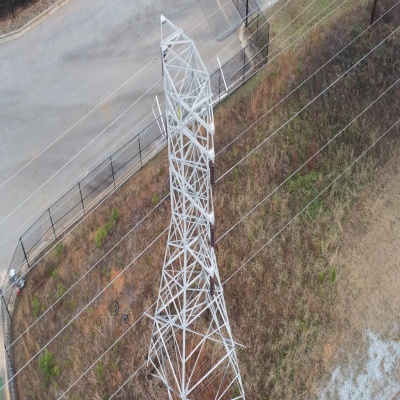}}\hfill
    \subfloat[]{\includegraphics[width=.24\linewidth]{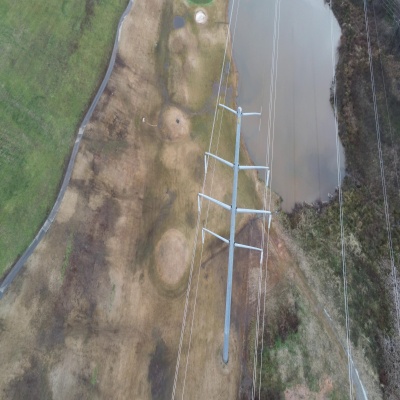}}
    \caption{Samples from the TTPLA dataset represent the main objects of categories (a) cable, (b) tower\_wooden, (c) tower\_lattice, (d) tower\_tucohy.}
    \label{fig:datasamples}
\end{figure}

\begin{figure}[h!]
    \centering
    \includegraphics[width=\linewidth]{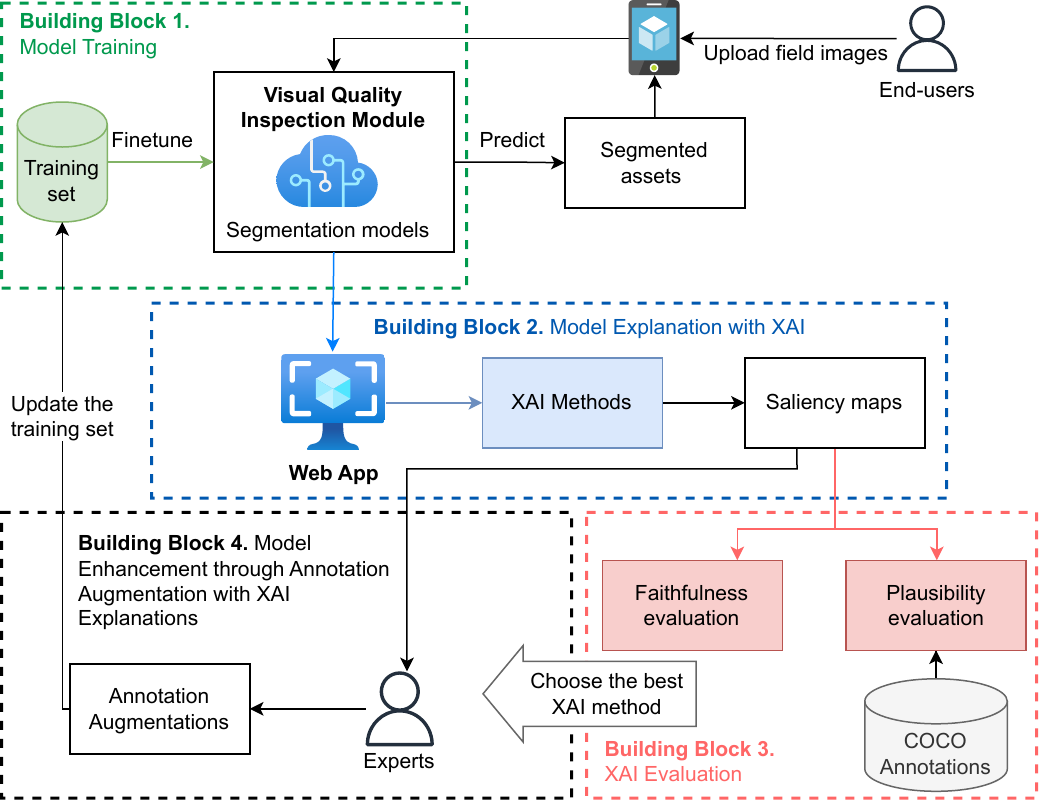}
    \caption{The enhanced \acf{VQI} framework integrated with XAI methods with 4 building blocks: (1) Training models, (2) Model Explanation with XAI, (3) XAI Evaluation, and (4) Model Improvement by XAI with Human-in-the-loop. The end-users interact with the framework via a web application.}
    \label{fig:evqi}
\end{figure}

\textbf{Enhanced \ac{VQI} Framework:} As illustrated in Fig.~\ref{fig:evqi}, our enhanced \ac{VQI} framework encompasses four pillars: semantic segmentation model training, XAI integration, XAI assessment, and model performance augmentation through XAI-guided annotations. Furthermore, we've built an interactive web application for easy access to the enhanced VQI framework.

\subsubsection{Model Training}\label{ss:training_model}
At this stage, the focal models are trained for the VQI module using a training subset from the original dataset. These images are resized, and their corresponding annotations are turned into masks for training purposes. We have chosen DeepLabv3-ResNet101 due to its mobile optimization and efficacy. The Dice loss function aids in training this model, providing an effective metric for the segmentation task at hand.

\subsubsection{Model Explanation with XAI}\label{ss:xai_inte}
Here, XAI methods extract explanation maps from the model's predictions. We harness several CAM-based XAI methods, known for their compatibility with semantic segmentation tasks. Through a web application, users can upload images and understand the model's rationale.

\subsubsection{XAI Evaluation}\label{ss:xai_eva}
This step assesses XAI techniques using plausibility and faithfulness criteria. By aligning explanations with human intuition and ensuring they mirror the model's logic, we can choose the most fitting XAI method for model enhancement.

\subsubsection{Model Enhancement via Annotation Augmentation with XAI Explanations}\label{ss:model_imp}
At this juncture, we amplify the performance of the DeepLabv3-ResNet101 model. Using data augmentation strategies and the best-performing XAI method from prior evaluations, we modify and enhance the dataset's annotations. After refining these annotations, the model is retrained, with the results from the original and improved models compared to validate the augmentation's efficacy. Lastly, this enhanced model is made available on mobile platforms.

\vspace{-12pt}

\section{Performance Evaluation}\label{sec:evaluation}
%
As stated in our contributions, this section details the results derived from our evaluation of CAM-based XAI techniques. Additionally, we discuss their use in improving model performance, specifically for applications on mobile devices.

\subsection{XAI Evaluation}\label{sec2}
%
\textbf{Evaluation Metrics:} We focus on two key metrics: plausibility and faithfulness of XAI explanations. 

Plausibility measures how explanations align with human understanding. Metrics used include:
  \begin{itemize}
      \item \textit{Energy-Based Pointing Game (EBPG)}: Evaluates precision and the XAI method's ability to highlight influential image regions~\cite{wang2020score}.
      \item \textit{\ac{IoU}}: Assesses localization and significance of attributions in the explanation map~\cite{chang2018explaining}.
      \item \textit{Bounding Box (Bbox)}: A variant of \ac{IoU} adapted to object size.
  \end{itemize}

Faithfulness evaluates how explanations match the model's decisions. Metrics include:
  \begin{itemize}
      \item \textit{Drop}: Measures the average decrease in model predictions using the explanation as input~\cite{fu2020axiom}.
      \item \textit{Increase}: Quantifies how often the model's confidence rises with the explanation as input~\cite{fu2020axiom}.
  \end{itemize}

\begin{figure}[h!]
\centering
    \subfloat[][Input image]{\includegraphics[width=.24\linewidth]{images/dataset/3_00092.jpg}}\hfill
    \subfloat[][Ground truth]{\includegraphics[width=.24\linewidth]{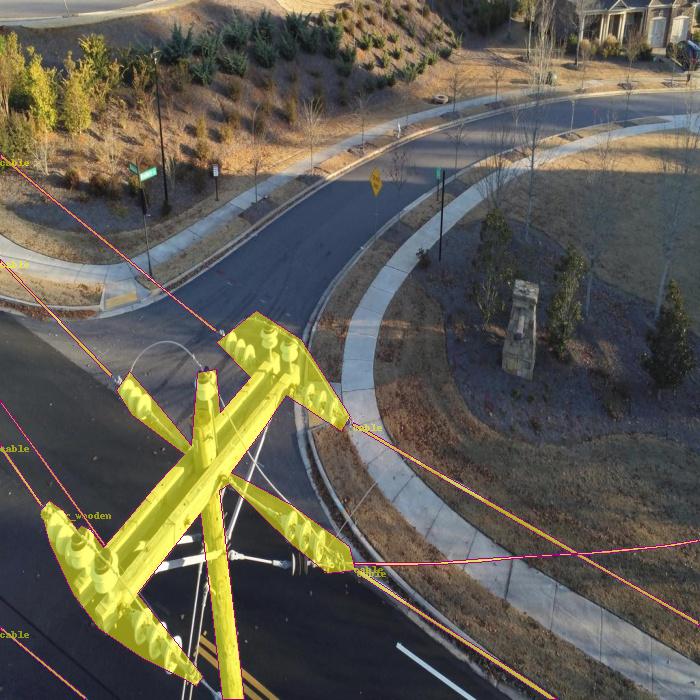}}\hfill
    \subfloat[][Segmentation]{\includegraphics[width=.24\linewidth]{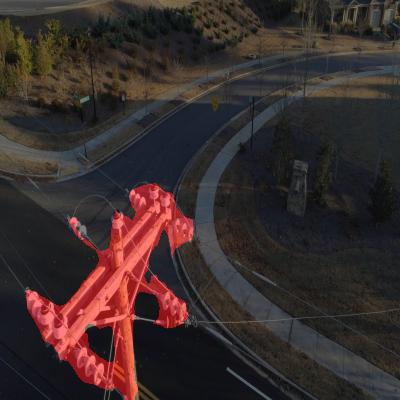}}\hfill
    \subfloat[][GradCAM]{\includegraphics[width=.24\linewidth]{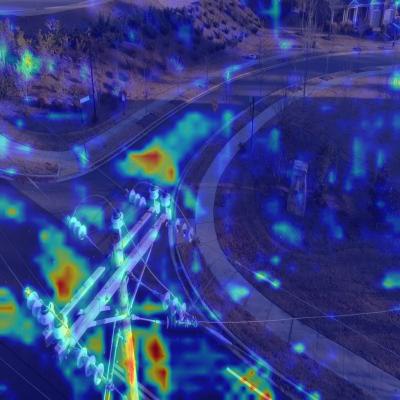}}\\
    \subfloat[][GradCAM++]{\includegraphics[width=.24\linewidth]{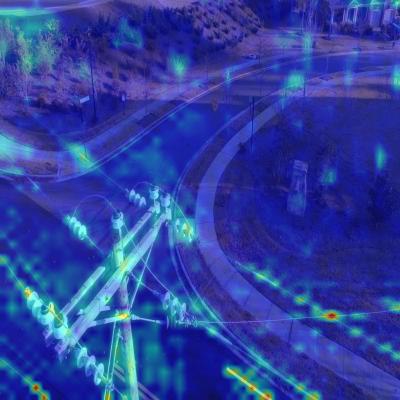}}\hfill
    \subfloat[][HiResCAM]{\includegraphics[width=.24\linewidth]{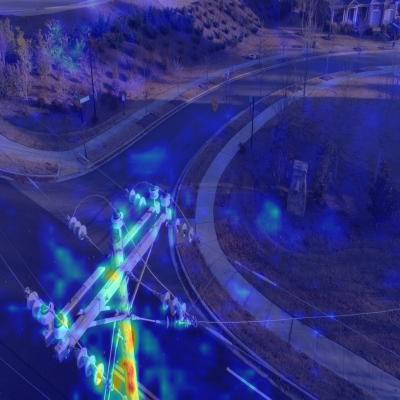}}\hfill
    \subfloat[][XGradCAM]{\includegraphics[width=.24\linewidth]{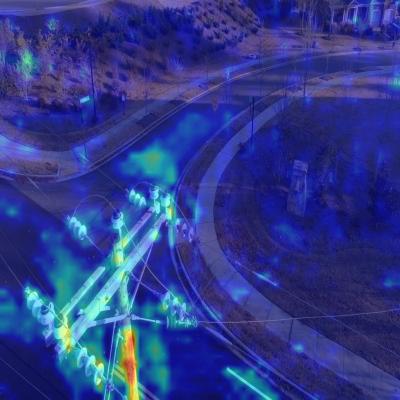}}\hfill
    \subfloat[][ScoreCAM]{\includegraphics[width=.24\linewidth]{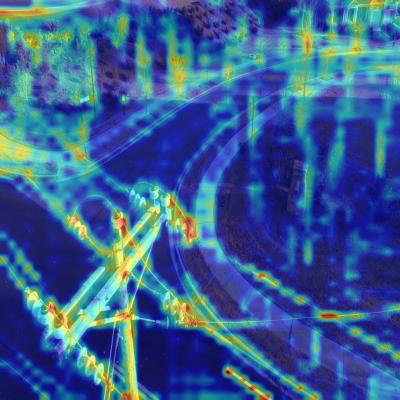}}\\
    \caption{The qualitative evaluation of implemented XAI methods on the segmentation result of the DeepLabv3-ResNet101 model on a sample from the test set. The category for the segmentation is the tower\_wooden denoted under the yellow box shown in the ground truth. The IoU value between the segmentation and the ground truth is 0.9085.}
    \label{fig:xai_images}
\end{figure}

\textbf{Evaluation Results:} The explanation maps of implemented XAI methods are demonstrated in Fig.~\ref{fig:xai_images}. The plausibility and faithfulness of XAI methods are quantitatively evaluated to find the most suitable XAI method, which can act as the core method of the model enhancement step. As shown in Table~\ref{tab:quantitative_result}, HiResCAM achieves not only the best performance in the faithfulness evaluations, such as Drop and Increase but also the shortest computational time. While GradCAM++ has the highest scores with BBox and IoU for plausibility, HiResCAM still performs plausibly with the highest score in EPBG. Hence, we choose HiResCAM as the core XAI method for the model enhancement step.

\begin{table}[h!]
\centering
\caption{The quantitative evaluations of XAI methods. For each metric, the arrow $\uparrow/\downarrow$ indicates higher/lower scores are better. The best is in bold.}\label{tab:quantitative}\label{tab:quantitative_result}
\resizebox{\linewidth}{!}{%
\begin{tabular}{lccccccc}
  \toprule
  Method & EPBG $\uparrow$ & BBox $\uparrow$ & IoU $\uparrow$ & Drop $\downarrow$ & Inc $\uparrow$ & Time(s) $\downarrow$ \\
  \midrule
  GradCAM & 50.49 & 48.39 & 47.94 & 5.21 & 52.57 & 3.21 \\
  GradCAM++ & 58.13 & \textbf{52.24} & \textbf{53.22} & 5.17 & 54.66 & 4.20 \\
  HiResCAM & \textbf{60.81} & 41.69 & 52.19 & \textbf{5.01} & \textbf{55.93} & \textbf{3.13} \\
  XGradCAM & 57.94 & 47.81 & 53.09 & 5.94 & 55.01 & 4.43 \\
  ScoreCAM & 54.01 & 43.95 & 51.94 & 7.34 & 47.19 & 52.50 \\
  \bottomrule
\end{tabular}}
\end{table}

\subsection{Model Enhancement} \label{sec3}
%
This section discusses the results of our attempt to enhance the DeepLabv3-ResNet101 model using XAI-guided annotation augmentation. Leveraging explanations generated by the XAI method for each training data sample, a domain expert skilled in semantic segmentation and XAI assists in refining the annotations. Using HiResCAM, we create explanations for various training samples.

\begin{figure}[h!]
    \centering
    \includegraphics[width=\linewidth]{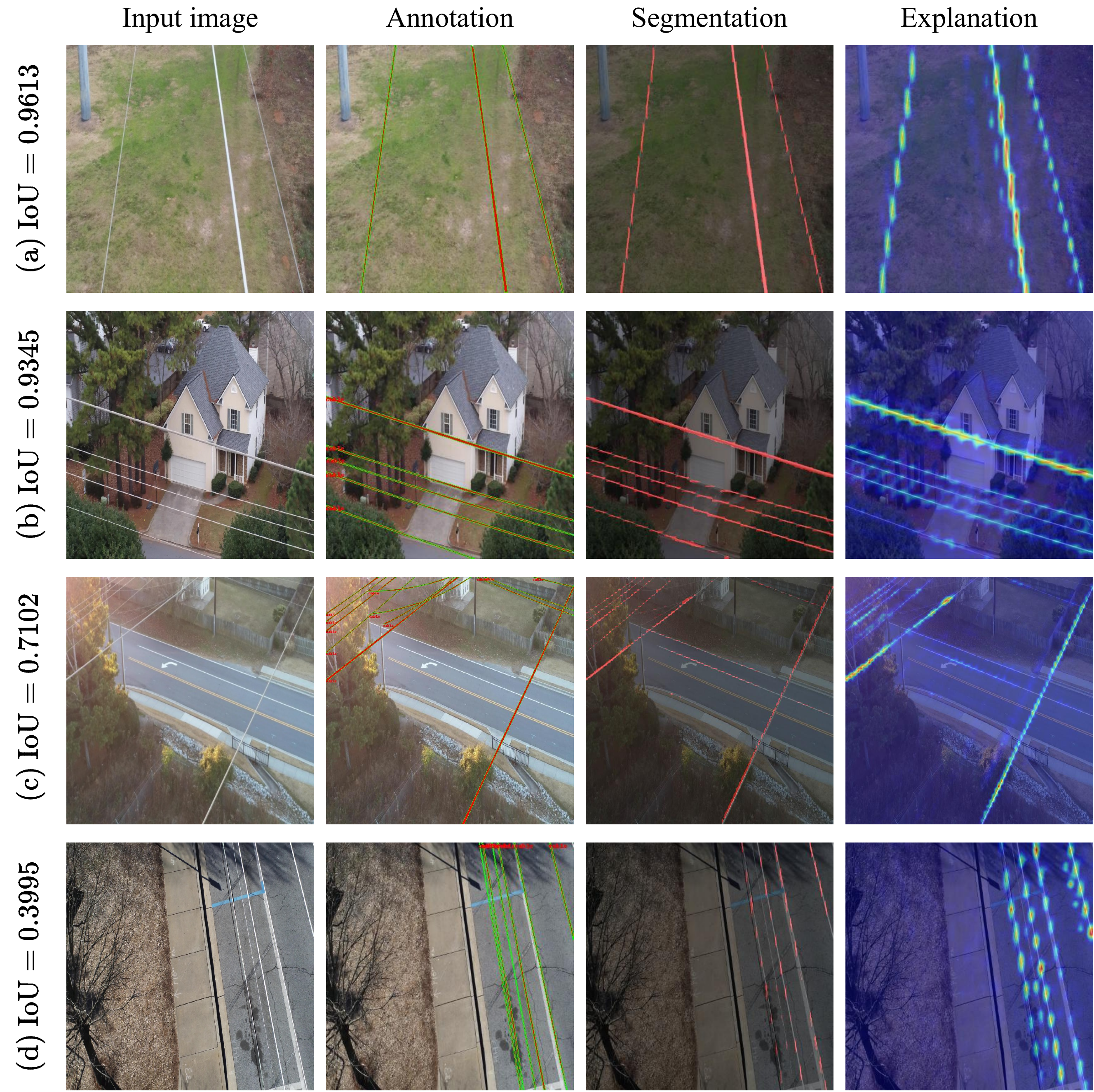}
    \caption{List of input images, COCO annotations (ground truth), segmentation results of the DeepLabv3-ResNet101 model, and the HiResCAM explanations in increasing order of complexity.}
    \label{fig:hirescamxaiimp}
\end{figure}

As evident in Fig.~\ref{fig:hirescamxaiimp}, the model excels in segmenting cables from simple backgrounds but struggles when similar objects are in the background. Explanations show that while the model focuses on the object and its immediate surroundings, it misses broader contextual cues in complex scenarios, a behavior attributed to models leveraging local and global context from initial annotations\cite{petsiuk2021black}.

\begin{figure}[h!]
    \centering
    \includegraphics[width=\linewidth]{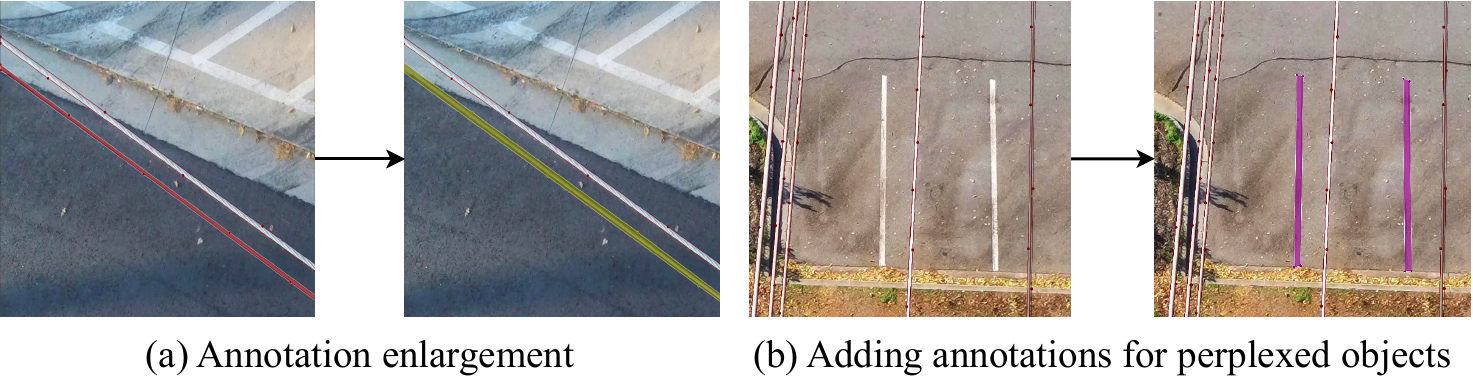}
    \caption{Annotation augmentation methods include: (a) Increasing annotation size for slender objects such as cables, and (b) Adding annotations for easily-confused elements, like road markings, to help the model differentiate them from objects like white cables.}
    \label{fig:fix_anno}
\end{figure}

\begin{figure}[h!]
    \centering
    \includegraphics[width=\linewidth]{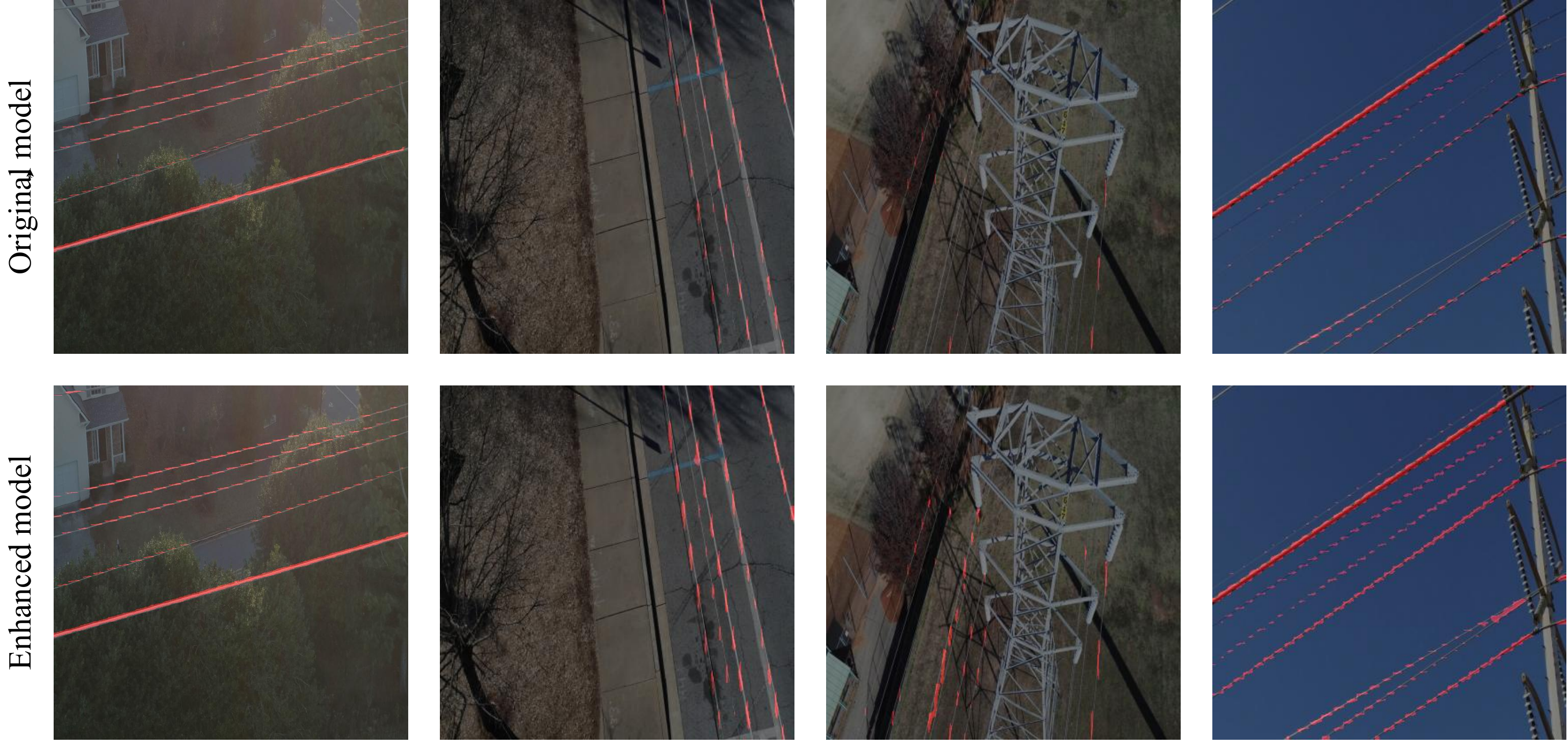}
    \caption{Qualitative results of DeepLabv3-ResNet101 before and after applying the enhancing model performance by annotation augmentation with XAI methods procedure.}
    \label{fig:qualitative_enhanced}
\end{figure}

To address this, the domain expert recommends two annotation augmentation strategies: \textit{Annotation Enlargement} and \textit{Adding Annotations for Perplexed Objects} (see Fig.~\ref{fig:fix_anno}). The improved model showcases enhanced segmentation abilities, as evident in Fig.~\ref{fig:qualitative_enhanced}. Notably, the enhanced model's IoU metrics, especially the $\mathtt{cable}$ IoU, improved significantly (from 55.06 to 58.11), leading to an overall IoU boost from 83.94 to 84.715, detailed in Table~\ref{tab:quantitatve_enhanced}.

\begin{table}[h!]
\centering
\caption{Quantitative results of DeepLabv3-ResNet101 before and after applying the enhancing model by annotation augmentation with XAI methods in IoU (\%) on each category and in average. The better is indicated in bold.}
\resizebox{\linewidth}{!}{%
\begin{tabular}{lccccc}
\toprule
Model & cable & tower\_wooden & tower\_lattice & tower\_tucohy & Overall \\
\midrule
Original & 55.06 & 94.75 & 95.31 & 90.63  & 83.94 \\
Enhanced & \textbf{58.11} & \textbf{94.78} & \textbf{95.32} & \textbf{90.65} & \textbf{84.715} \\
\bottomrule
\end{tabular}}

\label{tab:quantitatve_enhanced}
\end{table}

\section{Conclusion}\label{sec:conclusion}
%
This paper introduces an advanced \ac{VQI} system, integrating \ac{XAI} for improved interpretability and performance in mobile-based semantic segmentation. Using a public dataset, we demonstrated XAI's role in model enhancement. Multiple XAI methods were assessed, guiding users in choosing the most fitting techniques. Leveraging XAI explanations significantly bettered model results, especially in intricate scenarios. We aim to broaden our framework's application, targeting more image-related tasks. We also plan to refine the user interface, minimizing human intervention, and ensuring our approach's wider adaptability and diverse applicability.

\section*{Acknowledgment} 
This work was supported by the German Federal Ministry of Education and Research through grants 01IS17045, 02L19C155, 01IS21021A (ITEA project number 20219).
\balance

\bibliographystyle{IEEEtran}
\bibliography{IEEEabrv,ref}

\end{document}